\documentclass{ifacconf}

\usepackage{graphicx}      
\usepackage{natbib}    
\usepackage{mathtools}
\usepackage{amsmath}  
\usepackage{xcolor}
\usepackage{accents}
\usepackage{amsmath}
\usepackage{amssymb}
\usepackage{soul}
\usepackage[normalem]{ulem}

\begin{document}
\begin{frontmatter}

\title{Koopman Operator Based Time-Delay Embeddings and State History Augmented LQR for Periodic Hybrid Systems: Bouncing Pendulum and Bipedal Walking} 

\thanks[footnoteinfo]{ The work was supported by NSF grant 2128568}
\author{Chun-Ming Yang, Pranav A. Bhounsule}
\vspace{-0.25cm}

\address{Department of Mechanical and Industrial Engineering, \\ University of Illinois at Chicago, 
       842 W Taylor St, Chicago, IL 60607, USA. (e-mail: jyang241@uic.edu; pranav@uic.edu)
       }

\begin{abstract}  
Time-delay embedding is a technique that uses snapshots of state history over time to build a linear state space model of a nonlinear smooth system. We demonstrate that periodic non-smooth or hybrid system can also be modeled as a linear state space system using this approach as long as its behavior is consistent in modes and timings. We extend time-delay embeddings to generate a linear model of two periodic hybrid systems—the bouncing pendulum and the simplest walker—with control inputs. This leads to a  state history augmented linear quadratic regulator (LQR) which uses current and past state  history for feedback control. Example code can be found at \texttt{https://github.com/Chun-MingYang/koopman-timeDelay-lqr.git}
\end{abstract}


\end{frontmatter}

\section{Introduction}
Periodic hybrid systems are systems that exhibit both continuous dynamics  and discrete events, and do so in a repeating, cyclic manner. Some examples of periodic hybrid robotic systems include legged robots that walk, run, or climb; robotic arms or hands that brachiate, juggle, or swim; and manipulators that perform periodic pick-and-place tasks.  From a controls synthesis point of view, the hybrid nature of the dynamics are often tackled by using a switching controller but it is often difficult to  guarantee  the stability of such a controller.  Another approach involves using a two-layer controller \citep{bhounsule2012cornell,plooij2014open}. The first layer designs a nominal controller to generate a periodic solution. The second layer then linearizes the dynamics around this periodic solution and applies a linear controller to ensure stability. However, this method requires switching controllers at each discontinuity and is effective only for small perturbations near the periodic trajectory. Recently, the development of the Koopman operator has offered a new perspective for analyzing and controlling the nonlinear systems.


The Koopman operator \citep{koopman1931hamiltonian} is a linear operator that evolves a nonlinear dynamics linearly in high-dimensional function space. While the original formulation was limited to ergodic systems, \citep{mezic2005spectral} is the first work to extend the Koopman operator theory to general nonlinear systems, laying the foundation for later numerical methods—such as Dynamic Mode Decomposition (DMD) \citep{schmid2010dynamic, williams2015data, proctor2016dynamic}—that approximate the Koopman operator in a data-driven fashion. The Koopman operator has since become a powerful tool in system identification and control. So far, the Koopman operator has been increasingly applied to robotic systems, including soft robot \citep{bruder2020data}, quadcopter \citep{narayanan2023se}, underwater robot \citep{rahmani2024enhanced}, and quadrupedal legged robot \citep{yangkoopman, yang2025koopman}. However, most of the existing work has focused on continuous systems, with relatively few studies addressing hybrid systems.

The hybrid system poses greater challenges than the continuous system, in part because it is difficult to find suitable coordinates in high-dimensional space to capture its inherent transient features.  The earliest work \citep{govindarajan2016operator} demonstrates that Koopman analysis can be extended to hybrid systems and shows that the hybrid dynamics can be evolved in the associated eigenspaces.
Later, \citep{asada2023global} proposed a practical method—using direct encoding with radial basis functions—to encode hybrid dynamics into a global linear operator model in the Hilbert space.  Alternately, one can also use time-delay embedding to construct the high-dimensional space.





Time-delay embedding is a method used to lift the state space dimension by stacking a sequence of state history into a Hankel matrix based on measurements of generic observables—usually the state itself. The earliest work \citep{arbabi2017ergodic}  proved that one can find the Koopman operator by applying DMD to the time-delay embedded Hankel matrix, under the assumption of ergodic systems. Later, in the work of \citep{kamb2020time}, they established that projecting the time-delay coordinates onto an orthogonal basis yields a universal Koopman operator. (i.e., one invariant to the particular coordinate choice within the embedding.) This extension is grounded in Takens’ Embedding Theorem \citep{takens2006detecting}, which states that the time-delay embedding capture the structure of the original state space up to a diffeomorphism—meaning all topological and differential structures are preserved.



Taken’s Embedding Theorem, in its original form, does not apply to general non-smooth or hybrid systems. However, under certain specific conditions—such as when the switching behavior is periodic and consistent (i.e., following the same sequence of modes and timing)—the delay observables can preserve the hybrid features by embedding the switching non-smoothness within the periodic structure \citep{navarrete2019delay}.  Building on this insight, we apply the time-delay embedding approach to construct a Koopman model for the periodic hybrid system with control inputs—a contrast to past works \citep{haggerty2023control} that focused only on smooth systems. Once the model is obtained, we implement a linear quadratic regulator that leverages both current and past states for feedback control to stabilize the system.


\section{Methods}

\subsection{Koopman Operator for Systems with Control Inputs} 
The Koopman operator was originally presented as a method to create a global linear model of an uncontrolled system \citep{koopman1931hamiltonian}. The Koopman operator may be extended for a system with control inputs as follows. For a given non-linear system $\mathbf{x}_{i+1}=\mathbf{f}(\mathbf{x}_{i}, \mathbf{u}_{i})$, where $\mathbf{x}\in \mathbb{R}^{n}, \mathbf{u}\in \mathbb{R}^{m}$ are the system state and the control inputs respectively. Let ${\bf z}=[\mathbf{x}; \mathbf{u}] \in \mathbb{R}^{n+m}$, then there exists the observable function $\mathbf{g}({\bf z}): \mathbb{R}^{n+m}   \rightarrow \mathbb{R}^{p}$, such that the evolution of the system along this coordinate is characterized by a linear dynamics governed by an infinite dimensional Koopman operator $\mathcal{K}$ \citep{proctor2018generalizing} as: 
\begin{align} \label{eqn:koopman}
 \mathbf{g}(\mathbf{z}_{i+1})=\mathcal{K}\mathbf{g}(\mathbf{z}_{i})
\end{align}


A finite-dimensional approximation of the Koopman operator $\mathcal{K}$, denoted as $\mathbf{K}$, can be obtained using EDMD \citep{williams2015data}. Given $M+1$ snapshot (sample rate $dt$) pairs of state and input data, where $\mathbf{X}^{-} = [\mathbf{x}_{1}, \mathbf{x}_{2} \dots \mathbf{x}_{M}]$, $\mathbf{X}^{+} = [\mathbf{x}_{2}, \mathbf{x}_{3} \dots \mathbf{x}_{M+1}]$, $\mathbf{U}^{-} = [\mathbf{u}_{1}, \mathbf{u}_{2}  \dots \mathbf{u}_{M}]$ and $\mathbf{U}^{+} = [\mathbf{u}_{2}, \mathbf{u}_{3} \dots \mathbf{u}_{M+1}]$, we can write
\begin{align} \label{eqn:K}
    \mathbf{g}(\mathbf{X}^{+},\mathbf{U}^{+}) &= \mathbf{K} 
  \mathbf{g}(\mathbf{X}^{-}, \mathbf{U}^{-}) \nonumber \\
 \Rightarrow \mathbf{K} &= \mathbf{g}(\mathbf{X}^{+},\mathbf{U}^{+})  \mathbf{g}(\mathbf{X}^{-}, \mathbf{U}^{-})^{\dagger}
\end{align}
where $\bf{X}^{\dagger}$ is the pseudo-inverse of ${\bf X}$ and linear matrix $\mathbf{K} \in \mathbb{R}^{ p \times p}$ is computed via least-squares regression.


\subsection{Time-delay Embedding with Koopman Observables} \label{sec:tDMD}

The time-delay embeddings offer one technique for obtaining a linear dynamical system in the function space: We collect the history of states and control inputs and pass these into the observable ${\bf g}({\bf z}_i)$ and build the Hankel matrix, ${\bf H}_{i} \in{\mathbb{R}}^{p(N+1) \times p(M+1)}$ 
%
\begin{align}
    \mathbf{H}_i &= \left [ \begin{matrix}
    \mathbf{g}(\mathbf{z}_{i})  &    \mathbf{g}(\mathbf{z}_{i+1})&  \cdots&  \mathbf{g}(\mathbf{z}_{i+M})\\
    \mathbf{g}(\mathbf{z}_{i+1})  &    \mathbf{g}(\mathbf{z}_{i+2})&  \cdots&  \mathbf{g}(\mathbf{z}_{i+M+1})\\
              \vdots&           \vdots &  \ddots&  \vdots\\
    \mathbf{g}(\mathbf{z}_{i+N})  &  \mathbf{g}(\mathbf{z}_{i+N+1})&  \cdots&  \mathbf{g}(\mathbf{z}_{i+M+N})\\
\end{matrix} \right ] 
\end{align}
where $N$ and $M$ are the time embeddings with sample rate $dt$. We now use Koopman operator  given in Eqn.~\ref{eqn:K} to get
\begin{align} \label{eqn:hankel}
    \mathbf{H}_i &= \left [ \begin{matrix}
    \mathbf{g}(\mathbf{z}_{i})  &    \mathbf{K} \mathbf{g}(\mathbf{z}_{i})&  \cdots&  \mathbf{K}^M\mathbf{g}(\mathbf{z}_{i})\\
    \mathbf{K}\mathbf{g}(\mathbf{z}_{i})  &    \mathbf{K}^2\mathbf{g}(\mathbf{z}_{i})&  \cdots&  \mathbf{K}^{M+1}\mathbf{g}(\mathbf{z}_{i})\\
              \vdots&           \vdots &  \ddots&  \vdots\\
   \mathbf{K}^{N} \mathbf{g}(\mathbf{z}_{})  &  \mathbf{K}^{N+1}\mathbf{g}(\mathbf{z}_{i})&  \cdots&  \mathbf{K}^{M+N}\mathbf{g}(\mathbf{z}_{i})\\
\end{matrix} \right ] 
\end{align}
Since $\mathbf{K}$ is a linear operator, from Eqn. \ref{eqn:hankel} we can get
%
\begin{align} \label{eqn:L}
    \mathbf{H}_{i+1} = {\bf L} \mathbf{H}_{i} \Rightarrow \mathbf{L}= \mathbf{H}_{i+1} \mathbf{H}^{\dagger}_{i}
\end{align}
where $\bf{H}^{\dagger}$ is the pseudo-inverse of ${\bf H}$ and Koopman operator is $\mathbf{L} \in {\mathbb R}^{p(N+1)\times p(N+1)}$.


\vspace{-0.15cm}
\subsection{Time-delay Embedding Observables with State  and Control} \label{sec:tDMD}
\vspace{-0.2cm}

One common challenge in Koopman operator theory is that the finite-dimensional approximation of the Koopman matrix $\mathbf{K} \in \mathbb{R}^{p \times p}$ often leads to a high-dimensional system where $p \gg m + n$. Additionally, identifying a suitable high-dimensional observable space is both difficult and lacks theoretical guarantees. To address this, a practical approach is to use the identity function $\mathbf{g}(\mathbf{z}) = \mathbf{z} = [\mathbf{x}; \mathbf{u}]$, which has been empirically shown to work well with time-delay embeddings \citep{kamb2020time}. This approach allows for reducing the dimensionality of the Koopman operator $\mathbf{L}$, from $\mathbb{R}^{p(N+1) \times p(N+1)}$ to $\mathbb{R}^{(m+n)(N+1) \times (m+n)(N+1)}$, as shown in Eqn.~\ref{eqn:L}, provided the number of time-delay embeddings $N$ satisfies $(m + n)(N + 1) < p$, and simultaneously avoids the challenge of selecting appropriate observable functions.

We rewrite Eqn.~\ref{eqn:L} as follows
\begin{align} \label{linear_tdmd}
\begin{bmatrix}
 \mathbf{x}_{i+1}\\
 \mathbf{u}_{i+1}\\
 \mathbf{x}_{i+2}\\
 \mathbf{u}_{i+2}\\
 \vdots\\
 \mathbf{x}_{i+N+1}\\
 \mathbf{u}_{i+N+1}
\end{bmatrix} 
   = \mathbf{L} \begin{bmatrix}
 \mathbf{x}_{i}\\
 \mathbf{u}_{i}\\
 \mathbf{x}_{i+1}\\
 \mathbf{u}_{i+1}\\
 \vdots\\
 \mathbf{x}_{i+N}\\
 \mathbf{u}_{i+N}
\end{bmatrix} 
\Rightarrow 
\begin{bmatrix}
 \mathbf{x}_{i+1}\\
  \mathbf{x}_{i+2}\\
   \vdots\\
   \mathbf{x}_{i+N+1}\\
 \mathbf{u}_{i+1}\\
 \mathbf{u}_{i+2}\\
    \vdots\\
 \mathbf{u}_{i+N+1}
\end{bmatrix} 
   = \mathbf{\overline{L}} 
   \begin{bmatrix}
 \mathbf{x}_{i}\\
  \mathbf{x}_{i+1}\\
 \vdots\\
\mathbf{x}_{i+N}\\
 \mathbf{u}_{i}\\
 \mathbf{u}_{i+1}\\
 \vdots\\
 \mathbf{u}_{i+N}
\end{bmatrix}  
\end{align} 
where 
$\mathbf{\overline{L}} = {\bf P}{\bf L}\mathbf{P}^{T}$    and
$\mathbf{P} \in {\mathbb R}^{(m+n)(N+1)\times(m+n)(N+1)}$ is a permutation matrix that reorganizes the original basis structure into a form with all the states grouped at the top and the corresponding control inputs grouped at the bottom, and is defined as:
\begin{align}
   P_{i,j} = \begin{cases}
1, & \text{if } j = k(m+n) + r \text{ and } i = k\,n + r,\\
&k=0,\dots,N,\; r=1,\dots,n,\\
1, & \text{if } j = k(m+n) + n + s \text{ and } i = (N+1)n + \\
&k\,m + s, \quad k=0,\dots,N,\; s=1,\dots,m,\\
0, & \text{otherwise.}
\end{cases}
\end{align} 

\subsection{State History Augmented LQR Control} \label{sec:tdmd_lqr}
From Eqn.~\ref{linear_tdmd}, the state space model can be written as:
\begin{align} 
    \begin{bmatrix} 
    {\bf X}_{i+1} \\ {\bf U}_{i+1}  
    \end{bmatrix} 
    = 
    \begin{bmatrix} 
    \overline{\bf L}_{11} & \overline{\bf L}_{12}  \\ 
    \overline{\bf L}_{21} & \overline{\bf L}_{22}
    \end{bmatrix}
    \begin{bmatrix} 
    {\bf X}_{i} \\ {\bf U}_{i}  
    \end{bmatrix}  \label{eqn:state_space}
\end{align}
where ${\bf X}_{i} = [{\bf x}_i; {\bf x}_{i+1}\dots{\bf x}_{i+N}]$, ${\bf U}_{i} = [{\bf u}_i; {\bf u}_{i+1}\dots{\bf u}_{i+N}]$ and $\overline{\bf L}=\begin{bmatrix} 
    \overline{\bf L}_{11} & \overline{\bf L}_{12}  \\ 
    \overline{\bf L}_{21} & \overline{\bf L}_{22}
    \end{bmatrix}$ is a matrix with constants. Given Eqn.~\ref{eqn:state_space} and let the reference periodic trajectory be ${\bf X}^d$ and the corresponding control profile be ${\bf U}^d$,  the error dynamics can be formulated as:
\begin{equation} \label{eqn:error_dynamics}
    \begin{aligned}
    \begin{bmatrix} 
    {\bf X}_{i+1}-{\bf X}^d_{i+1} \\ {\bf U}_{i+1} -{\bf U}^d_{i+1} 
    \end{bmatrix} 
    &= 
    \begin{bmatrix} 
    \overline{\bf L}_{11} & \overline{\bf L}_{12}  \\ 
    \overline{\bf L}_{21} & \overline{\bf L}_{22}
    \end{bmatrix}
    \begin{bmatrix} 
    {\bf X}_{i}-{\bf X}^d_{i} \\ {\bf U}_{i} - {\bf U}^d_{i} 
    \end{bmatrix}    \\
    \begin{bmatrix}
        \mathbf{\hat{X}}_{i+1}\\
        \mathbf{\hat{U}}_{i+1}
    \end{bmatrix}&=\begin{bmatrix} 
    \overline{\bf L}_{11} & \overline{\bf L}_{12}  \\ 
    \overline{\bf L}_{21} & \overline{\bf L}_{22}
    \end{bmatrix}\begin{bmatrix}
        \mathbf{\hat{X}}_{i}\\
        \mathbf{\hat{U}}_{i}
    \end{bmatrix}
    \end{aligned}
\end{equation}
%

where ${\bf \hat{X}_{i}} = {\bf X}_{i}-{\bf X}^d_{i}$ and ${\bf \hat{U}_{i}} = {\bf U}_{i}-{\bf U}^d_{i}$, then the first  $n \times (N+1)$ rows of left hand side of Eqn. \ref{eqn:error_dynamics} can be written as: 
\begin{equation}
\begin{aligned}
    \mathbf{\hat{X}}_{i+1}
    &=  
    \overline{\bf L}_{11}\mathbf{\hat{X}}_{i}
    + 
    \overline{\bf L}_{12}   \hat{\bf U}_i \\
    &= 
    {\bf A}  \mathbf{\hat{X}}_{i} + {\bf B} \hat{\bf U}_i
\end{aligned}
\end{equation}

where ${\bf A} \in {\mathbb R}^{n(N+1) \times n(N+1)}$ and ${\bf B} \in {\mathbb R}^{n(N+1) \times m(N+1)}$ are the system matrices. Consider using controller ${\bf U}_{i} =  {\bf U}^d_{i} + \hat{\bf U}_i$, we can now use the linear quadratic regulator (LQR) to compute a gain ${\bf K}_{\mbox{\scriptsize LQR}}  \in {\mathbb R}^{m(N+1) \times n(N+1)}$ such that the optimal correction control term $\hat{\bf U}_i = -{\bf K}_{\mbox{\scriptsize LQR}}{\bf \hat{X}_{i}}$ can be retrieved. It is important to note that unlike traditional LQR which uses the current state for feedback control, the controller uses the current state and past state history for feedback control. This ensures that the controller is retrospective and less reactive.

\subsection{Application to Periodic Hybrid Systems}

The Takens Embedding Theorem \citep{takens2006detecting}  states that for a smooth deterministic, non-noisy dynamical system, the time-delay embeddings map (see Sec.~\ref{sec:tDMD}) is a diffeomorphism (i.e., a smooth, invertible mapping) of the original state space provided rich trajectory data are used for the mapping. In particular, if $d$ is the number of delays and $n$ is the dimension of state space then $d>2n+1$.

Although the Takens Embedding Theorem applies only to smooth systems, it can be applied to periodic hybrid systems as long as, $d$, the number of delays, encompasses one periodic cycle of the system. This can be intuitively understood as follows. When the trajectory data is periodic, it can be represented using a Fourier Series, in other words, it is smooth. Hence the Takens Embedding Theorem could be applied. However, any mode and/or timing changes in the switching will make the system non-smooth and violate the Takens Embedding Theorem.

We present the dynamics of two hybrid systems: a hybrid pendulum bouncing model and a planar model of bipedal walking known as the simplest walker.

\begin{figure}
\begin{center}
\includegraphics[scale=0.9]{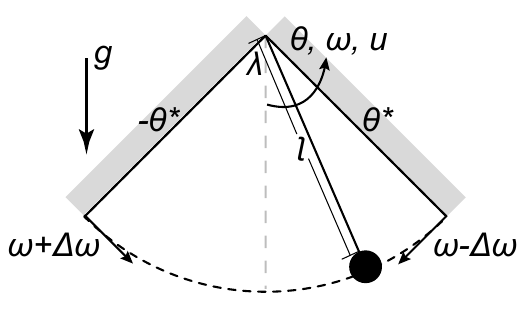}    
\caption{Bouncing pendulum model \citep{govindarajan2016operator}} 
\label{fig:hybrid_pendulum}
\end{center}
\end{figure}

\subsubsection{A. Hybrid Pendulum Model:}
The hybrid pendulum model, shown in Fig.~\ref{fig:hybrid_pendulum} , exhibits a periodic hybrid dynamics—It consists of a swing phase represented by a continuous dynamics $\mathbf{\dot{x}}=\mathbf{f}(\mathbf{x}, \mathbf{u})$ and  impact phases represented by  discrete mappings with switch conditions  $\mathbf{x}^{+}=\begin{Bmatrix} \mathcal{R}_{i}(\mathbf{x})\:|\: \mathcal{S}_{i}(\mathbf{x})=0 \end{Bmatrix}$—with the periodic limit cycle $t \in [0, T]$ as:
\begin{align}
    \mathbf{x}&=\mathbf{x}_{0} \\
    \mathbf{\dot{x}}&=\mathbf{f}(\mathbf{x}, u)
\end{align}
\begin{equation}
    \mathbf{x}^{+}= \begin{Bmatrix} \mathcal{R}_{i}(\mathbf{x})\:|\: \mathcal{S}_{i}(\mathbf{x})=0 \end{Bmatrix} , \:\text{for}\:i=1,2
\end{equation}
where $\mathbf{x}, u$ are the dynamics state and non-dimensional control torque respectively. By defining the state vector $\mathbf{x}=[\theta, \omega]^{T}$, the continuous dynamics  can be written as:
\begin{equation}
    \mathbf{\dot{x}}=\mathbf{f}(\mathbf{x}, \mathbf{u})=\begin{bmatrix}
 \omega\\
 \frac{-g}{l}\sin\theta - \lambda\omega+u
\end{bmatrix}
\end{equation}
where $g,l,\lambda$ are gravity term, pendulum length, and damping constant respectively. The switching conditions $\mathcal{S}_{i}(\mathbf{x})=0$ are defined when pendulum passing through the given angles $\pm \theta^{*}$ with the discrete reset map $\mathcal{R}_{i}(\mathbf{x})$—an instantaneous opposite kick—defined as:
\begin{equation}
\begin{aligned}
\mathcal{S}_{1}(\mathbf{x}) : \quad & \Big\{ \theta^{-} + \theta^{*} \,\big|\, \omega^{-} < 0 \Big\} = 0, \\
\mathcal{R}_{1}(\mathbf{x}) : \quad & 
\begin{cases}
\theta^{+} = -\theta^{*},\\[1mm]
\omega^{+} = \omega^{-} + \Delta\omega,
\end{cases} \\
\mathcal{S}_{2}(\mathbf{x}) : \quad & \Big\{ \theta^{-} - \theta^{*} \,\big|\, \omega^{-} > 0 \Big\} = 0, \\
\mathcal{R}_{2}(\mathbf{x}) : \quad & 
\begin{cases}
\theta^{+} = \theta^{*},\\[1mm]
\omega^{+} = \omega^{-} - \Delta\omega.
\end{cases}
\end{aligned}
\end{equation}

where $-,+$ are the notations representing the instance just before and just after the switch conditions; $\Delta\omega>0$ is the instantaneous kick modeled as instant augular velocity change. The limit cycle of the system can be obtained by considering \(u=\lambda\omega\)—which turns the system into an undamped pendulum that is naturally periodic—such that the periodic orbit, given \(\mathbf{x}_{0}=[\theta_{0}, \omega_{0}]^T\), can be obtained manually by selecting $\theta^{*}=\begin{Bmatrix} \theta(t_{s})| t_{s}\in[0, T]\end{Bmatrix}<max(\theta(0:T))$ and $\Delta\omega=2|\omega(t_{s})|$.

\begin{figure*}
\begin{center}
\includegraphics[width=16.5cm]{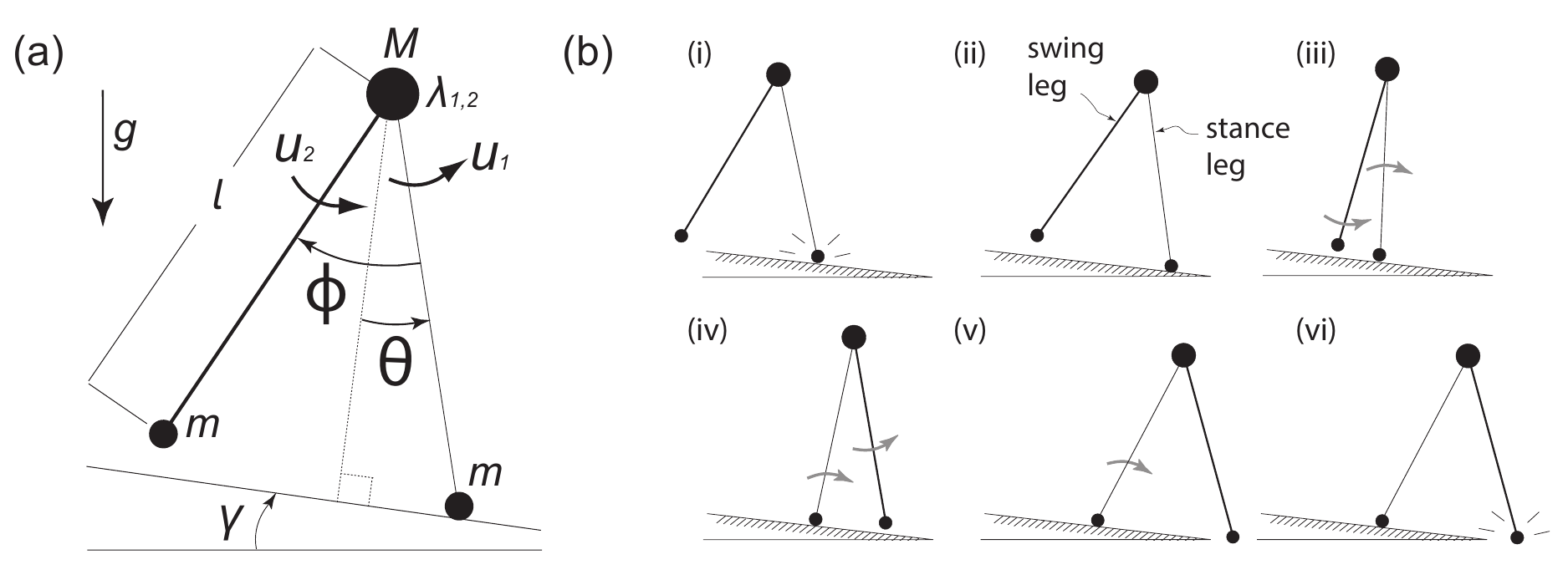}    
\caption{(a) Simplest walker \citep{garcia1998simplest}, note that the model has the damping terms and control torques on hips of both stance and swing legs. (b) A typical step of the simplest walker.} 
\label{fig:walker_model}
\end{center}
\end{figure*}

\subsubsection{B. Bipedal Walking Model:}
The simplified walking model, shown in Fig.~\ref{fig:walker_model}, exhibits periodic hybrid dynamics—It consists of a single stance phase represented by continuous dynamics $\dot{\mathbf{x}}=\mathbf{f}(\mathbf{x},\mathbf{u})$ and a support exchange phase represented by discrete mapping $\mathbf{x}^{+}=\mathcal{R}(\mathbf{x}(t))$—with the periodic limit cycle $t\in[0,T]$ governed by switch condition $S(\mathbf{x}|t=T)=0$ as:
\begin{align} \label{hybrid_dynamics}
        t=0&:\quad \mathbf{x}=\mathbf{x}_{0}\\
        0 \leq t \leq T&:\quad \dot{\mathbf{x}}=\mathbf{f}(\mathbf{x},\mathbf{u})\\
        t=T&:\quad S(\mathbf{x}(t))=0 \\
        t=T&:\quad\mathbf{x}^{+}=\mathcal{R}(\mathbf{x}(t))
\end{align}
where $\mathbf{x}=[\theta, \phi, \dot{\theta}, \dot{\phi}]^{T}$ is the dynamics state vector with $\theta, \dot{\theta}$ for stance leg angle and angular velocity and $\phi, \dot{\phi}$ for swing foot respectively; $\mathbf{u}=[u_{1}, u_{2}]^{T}$ is the control vector consisted with the stance and swing leg hip torques respectively. 

By taking moments about stance foot contact point and hip hinge respectively, and non-dimensionalizaing time with $ \sqrt{l/g} $, and applying the limit $m/M \rightarrow0$, and non-dimensionalizing torque by $ Mgl $, we obtain the continuous dynamics during single stance phase as: \begin{align} \label{single_stance}
    \ddot{\theta}&=\sin(\theta-\gamma) - \lambda_1\dot{\theta} + u_{1}\\
    \ddot{\phi}&=\sin(\theta-\gamma) + \left\{ \dot{\theta}^{2} - \cos(\theta-\gamma) \right\}\sin(\phi) -\lambda_2\dot{\phi}+u_{2}
\end{align}
where $M,m,l$ are mass of hip, mass and length of the foot respectively; $\gamma$ is the slope of the ramp; $\lambda_i$ ($i=1,2$) are the damping coefficient. The switch condition $S(\mathbf{x}|t=T)$ is when the swing foot collides with the ramp but ignoring the condition when the legs are parallel, i.e., $\phi=\theta=0$, followed by support exchange phase where the legs exchange roles, $\mathbf{x}^{+}=\left\{\mathcal{R}(\mathbf{x})|t=T \right\} $ as:
\begin{equation} \label{foot_strik}
    S(\mathbf{x}|t=T):\phi^{-}-2\theta^{-}=0
\end{equation}
\begin{equation} \label{discrete_mapping}
    \mathcal{R}(\mathbf{x}|t=T):
    \left\{\begin{matrix}
    \begin{aligned}
   \theta^{+}&=-\theta^{-}\\
   \phi^{+}&=-\phi^{-}=-2\theta^{-}\\
   \dot{\theta}^{+}&=\cos(2\theta^{-})\dot{\theta}^{-}\\
   \dot{\phi}^{+}&=\left\{ 1-\cos(2\theta^{-})\right\}\cos(2\theta^{-})\dot{\theta}^{-}
    \end{aligned}
\end{matrix}\right.
\end{equation}
where $-,+$ are the notations representing the instance just before and just after the foot strike event.  The fixed points 
can be found by solving the following Poincaré analysis as:
\begin{align}
        \mathcal{P}:\begin{Bmatrix}
        (\mathbf{x},t)|t=0
        \end{Bmatrix}&\rightarrow{}\begin{Bmatrix}
            (\mathbf{x},t)|t=T^{*} 
        \end{Bmatrix} \\
        \mathbf{x}^{*}=\mathcal{P}(&\mathbf{x}^{*},T^{*})
\end{align}
\begin{figure}[t]
\begin{center}
\includegraphics[scale=0.9]{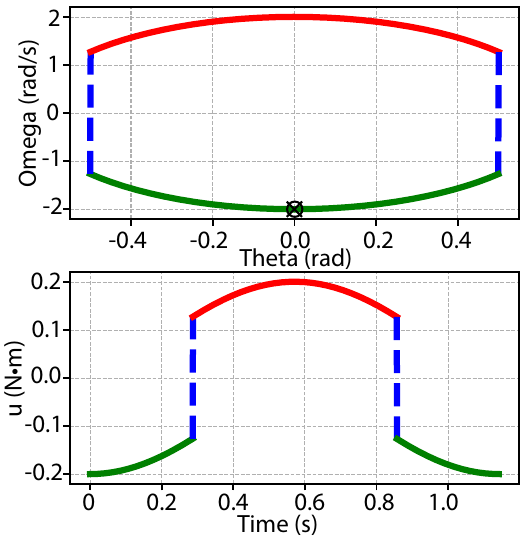}    
\caption{Bouncing pendulum limit cycle. The red and green lines respectively represent positive and negative rotation direction, and blue lines represent the hybrid switch.} 
\label{fig:pd_orbit}
\end{center}
\end{figure}
\vspace{-0.25cm}
\begin{figure}
\begin{center}
\includegraphics[scale=0.9]{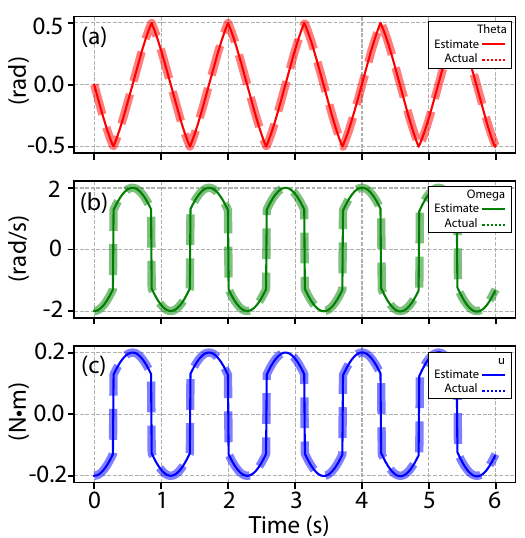}    
\caption{Bouncing pendulum Koopman prediction.} 
\label{fig:bound_est}
\end{center}
\end{figure}
\begin{figure}
\begin{center}
\includegraphics[scale=0.9]{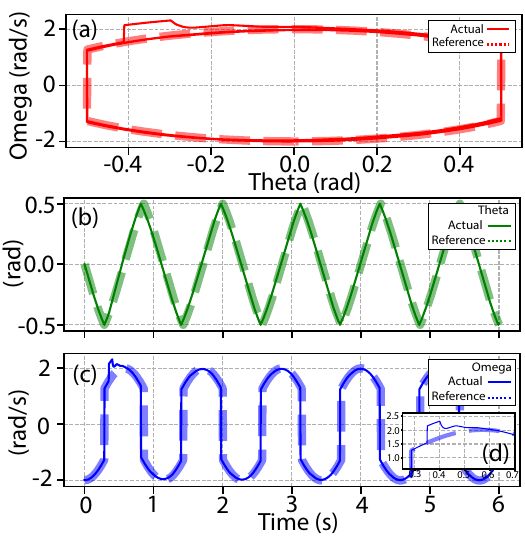}    
\caption{Bouncing pendulum LQR control.} 
\label{fig:bound_lqr}
\end{center}
\end{figure}

 

\section{Results}

\begin{figure}
\begin{center}
\includegraphics[scale=0.9]{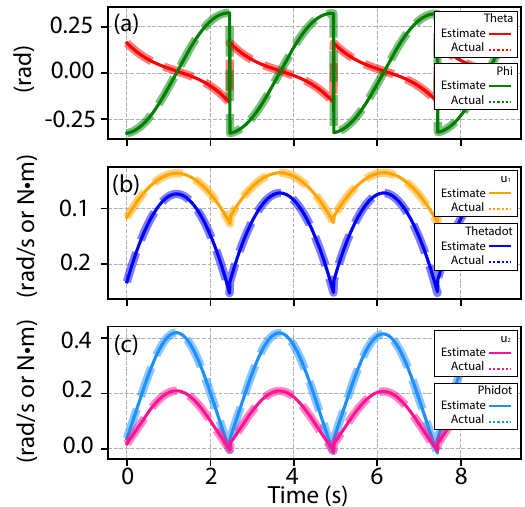}    
\caption{Simplest walker Koopman prediction. } 
\label{fig:walker_est}
\end{center}
\end{figure}

\begin{figure}[t]
\begin{center}
\includegraphics[scale=0.9]{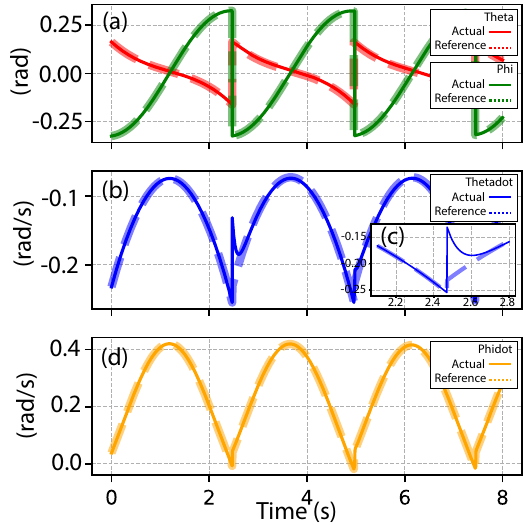}    
\caption{Simplest walker LQR control.} 
\label{fig:walker_lqr}
\end{center}
\end{figure}

\subsection{Bouncing Pendulum}
For pendulum with length $l=1$ m, damping constant $\lambda=0.1$, the limit cycle shown in Fig.~\ref{fig:pd_orbit}  corresponding to the initial position $\theta_{0}=0$ rad, $\omega_{0}=-2$ rad/s can be obtained by selecting $\theta^{*}=0.5$ rad and $\Delta\omega=2.538$ rad/s, resulting the period to be $T=1.144$ sec, such that the reference trajectory and nominal control profile, which also serve as the training dataset, are collected by simulating the model, initialed with Poincaré fixed points, for $6$ sec, as shown as dashed lines in Fig.~\ref{fig:bound_est}.

To obtain the linear model for hybrid pendulum dynamics, the Hankel matrix is formulated with the parameters $N=110, M=90$ and  $dt=1e^{-2}$ as the sample rate, then the Koopman operator $\bar{\mathbf{L}}$ is approximated from EDMD as discussed in Sec.\ref{sec:tDMD}. The resulting linear model demonstrates accurate prediction, as shown as solid lines shown in Fig.~\ref{fig:bound_est}, with root mean square errors (RMSE) of $0.008, 0.015$, and $0.017$ for $\theta$, $\omega$, and $u$, respectively.

The LQR controller is used to handle the external disturbance---an impulse $\omega^{+}(t=0.35)=\omega^{-}(t=0.35)+0.6$ applied to pendulum at time $0.35$ sec. So as to calculate the feedback gain, the system matrices $\mathbf{A}$ and $\mathbf{B}$ are first extracted from $\bar{\mathbf{L}}$ as described in Sec.~\ref{sec:tdmd_lqr}. The MATLAB  function \texttt{d2c()} is then used to convert them to their continuous-time forms, $\mathbf{A}_c$ and $\mathbf{B}_c$. Finally, the LQR gain $\mathbf{K}_{\text{LQR}}$ is computed using the MATLAB function \texttt{lqr()}. One can see that with the LQR controller, the periodic trajectories can be stabilized, shown in Fig. \ref{fig:bound_lqr}, with the RMSE of $0.035, 0.496$ for tracking errors to $\theta, \omega$ respectively.

\subsection{Simplest Walker}

Given the initial range $\theta_{0}\in[0.1,0.2],\phi_{0}\in[-0.4,-0.3]$ rad and $\dot{\theta}_{0}\in[-0.3,-0.2],\dot{\phi}_{0} \in [0.03,0.04] $ rad/s, along with the nominal controller $u_{1}=\lambda_1\dot{\theta},u_{2}=\lambda_2\dot{\phi}$ ($\lambda_1 = \lambda_2 =0.5$) to cancel the damping effect, the fixed points of the periodic walking limit cycle—whose Poincaré section is taken right after the foot-strike event—are found through Poincaré analysis to be $[\theta_{0},\phi_{0}, \dot{\theta}_{0}, \dot{\phi}_{0}]^{*} = [0.162, -0.325, -0.231, 0.038]$ rad, such that the reference trajectory and norminal control profile, which also serve as the training dataset, are collected by simulating the model, initialed with Poincaré fixed points, for $8$ sec, as shown as dashed lines in Fig.~\ref{fig:walker_est}.

To obtain the linear model for simplest walker, the Hankel matrix is formulated with the parameters $N=300, M=600$ and  $dt=1e^{-2}$ as the sample rate, then the Koopman operator $\bar{\mathbf{L}}$ is approximated from EDMD as discussed in Sec.\ref{sec:tDMD}. The resulting linear model demonstrates accurate prediction, as shown as solid lines shown in Fig.~\ref{fig:walker_est}, with root mean square errors (RMSE) of $0.001, 0.002, 0.001, 0.003, 0.011$, and $0.006$ for $\theta$, $\phi$, $\dot{\theta}$, $\dot{\phi}$, $u_{1}$, and $u_{2}$ respectively.

In order to handle the external disturbance---an impulse applied to the stance foot just after the second foot-strike---an LQR controller is used. So as to calculate the feedback gain, the system matrices $\mathbf{A}$ and $\mathbf{B}$ are first extracted from $\bar{\mathbf{L}}$ as described in Sec.~\ref{sec:tdmd_lqr}. The MATLAB function \texttt{d2c()} is then used to convert them to their continuous-time forms, $\mathbf{A}_c$ and $\mathbf{B}_c$. Finally, the LQR gain $\mathbf{K}_{\text{LQR}}$ is computed using the MATLAB function \texttt{lqr()}. One can see that with the LQR controller, the periodic trajectories can be stabilized, shown in Fig. \ref{fig:walker_lqr}, with the RMSE of $0.019, 0.038, 0.006, 0.001 $ for tracking errors to $\theta, \phi, \dot{\theta}, \dot{\phi}$ respectively.

\section{Discussion and Conclusion}
This paper demonstrates that a sequence of state and control snapsnots over time -- known as time-delay embeddings -- can be used to compute a linear representation of the dynamics of a periodic hybrid system. The resulting linear dynamics can be used with a standard LQR controller for feedback control. The method is demonstrated on two hybrid systems: a pendulum that bounces on a wall and the simplest walker.

The time delay embedding like the Koopman operator can create a global linear state space model by projecting on high-dimensional function space \citep{kamb2020time,brunton2017chaos}. Koopman operator relies on a good choice of observables that span the dynamics of the system. This is often done by trial and error and there is no guarantee of finding such a set of observables. On the other hand, time embedding with system states as observable is able to faithfully reconstruct the dynamics, which is a major advantage of time delay embeddings. Thus Koopman operator lifts using non-linear functions (observers) while time embedding lifts using a set of delay coordinates of the state space thus turning the temporal dynamics into spatial representation.


The LQR controller resulting from time-delay embedding is the state history augmented control which uses the current and past history for feedback control. Unlike traditional LQR which is reactive or MPC which is pro-active, the state history augmented LQR is retrospective. It potentially leads to a smoother control than a purely reactive controller as it takes into account past state history.

Our work has some limitations. In order to apply time embedding to hybrid systems without violating the Takens Embedding Theorem, we had to restrict periodic hybrid systems where the guards, events, and mode changes are unaffected by the disturbance. Consequently, we can only control fully actuated periodic hybrid systems as full actuation provides enough control authority to ensure the system can maintain periodicity in the presence of a disturbance. Moreover, the resulting linear lifting was local to the periodic hybrid system that it modeled. Finally, the resulting linearized system was of high dimension due to use of sufficiently long time history. This can be remediated by using model reduction techniques and is left as a future work. 

We conclude that time delay embeddings provides a powerful method for modeling periodic hybrid systems as long as the systems are consistent. That is, the modes and timings remains unaffected by control or disturbance. 

\bibliography{ifacconf}          

\begin{thebibliography}{22}
\providecommand{\natexlab}[1]{#1}
\providecommand{\url}[1]{\texttt{#1}}
\providecommand{\urlprefix}{URL }
\expandafter\ifx\csname urlstyle\endcsname\relax
  \providecommand{\doi}[1]{doi:\discretionary{}{}{}#1}\else
  \providecommand{\doi}{doi:\discretionary{}{}{}\begingroup \urlstyle{rm}\Url}\fi

\bibitem[{Arbabi and Mezic(2017)}]{arbabi2017ergodic}
Arbabi, H. and Mezic, I. (2017).
\newblock Ergodic theory, dynamic mode decomposition, and computation of spectral properties of the koopman operator.
\newblock \emph{SIAM Journal on Applied Dynamical Systems}, 16(4), 2096--2126.

\bibitem[{Asada(2023)}]{asada2023global}
Asada, H.H. (2023).
\newblock Global, unified representation of heterogenous robot dynamics using composition operators: A koopman direct encoding method.
\newblock \emph{IEEE/ASME Transactions on Mechatronics}, 28(5), 2633--2644.

\bibitem[{Bhounsule et~al.(2012)Bhounsule, Cortell, and Ruina}]{bhounsule2012cornell}
Bhounsule, P., Cortell, J., and Ruina, A. (2012).
\newblock Cornell ranger: Implementing energy-optimal trajectory control using low information, reflex-based control.
\newblock \emph{Dyn Walk. Pensacola}.

\bibitem[{Bruder et~al.(2020)Bruder, Fu, Gillespie, Remy, and Vasudevan}]{bruder2020data}
Bruder, D., Fu, X., Gillespie, R.B., Remy, C.D., and Vasudevan, R. (2020).
\newblock Data-driven control of soft robots using koopman operator theory.
\newblock \emph{IEEE Transactions on Robotics}, 37(3), 948--961.

\bibitem[{Brunton et~al.(2017)Brunton, Brunton, Proctor, Kaiser, and Kutz}]{brunton2017chaos}
Brunton, S.L., Brunton, B.W., Proctor, J.L., Kaiser, E., and Kutz, J.N. (2017).
\newblock Chaos as an intermittently forced linear system.
\newblock \emph{Nature communications}, 8(1), 19.

\bibitem[{Garcia et~al.(1998)Garcia, Chatterjee, Ruina, and Coleman}]{garcia1998simplest}
Garcia, M., Chatterjee, A., Ruina, A., and Coleman, M. (1998).
\newblock The simplest walking model: stability, complexity, and scaling.

\bibitem[{Govindarajan et~al.(2016)Govindarajan, Arbabi, Van~Blargian, Matchen, Tegling et~al.}]{govindarajan2016operator}
Govindarajan, N., Arbabi, H., Van~Blargian, L., Matchen, T., Tegling, E., et~al. (2016).
\newblock An operator-theoretic viewpoint to non-smooth dynamical systems: Koopman analysis of a hybrid pendulum.
\newblock In \emph{2016 IEEE 55th Conference on Decision and Control (CDC)}, 6477--6484. IEEE.

\bibitem[{Haggerty et~al.(2023)Haggerty, Banks, Kamenar, Cao, Curtis, Mezi{\'c}, and Hawkes}]{haggerty2023control}
Haggerty, D.A., Banks, M.J., Kamenar, E., Cao, A.B., Curtis, P.C., Mezi{\'c}, I., and Hawkes, E.W. (2023).
\newblock Control of soft robots with inertial dynamics.
\newblock \emph{Science robotics}, 8(81), eadd6864.

\bibitem[{Kamb et~al.(2020)Kamb, Kaiser, Brunton, and Kutz}]{kamb2020time}
Kamb, M., Kaiser, E., Brunton, S.L., and Kutz, J.N. (2020).
\newblock Time-delay observables for koopman: Theory and applications.
\newblock \emph{SIAM Journal on Applied Dynamical Systems}, 19(2), 886--917.

\bibitem[{Koopman(1931)}]{koopman1931hamiltonian}
Koopman, B.O. (1931).
\newblock Hamiltonian systems and transformation in hilbert space.
\newblock \emph{Proceedings of the National Academy of Sciences}, 17(5), 315--318.

\bibitem[{Mezi{\'c}(2005)}]{mezic2005spectral}
Mezi{\'c}, I. (2005).
\newblock Spectral properties of dynamical systems, model reduction and decompositions.
\newblock \emph{Nonlinear Dynamics}, 41(1), 309--325.

\bibitem[{Narayanan et~al.(2023)Narayanan, Tellez-Castro, Sutavani, and Vaidya}]{narayanan2023se}
Narayanan, S.S., Tellez-Castro, D., Sutavani, S., and Vaidya, U. (2023).
\newblock Se (3) koopman-mpc: Data-driven learning and control of quadrotor uavs.
\newblock \emph{IFAC-PapersOnLine}, 56(3), 607--612.

\bibitem[{Navarrete and Viswanath(2019)}]{navarrete2019delay}
Navarrete, R. and Viswanath, D. (2019).
\newblock Delay embedding of periodic orbits using a fixed observation function.
\newblock \emph{Physica D: Nonlinear Phenomena}, 388, 1--9.

\bibitem[{Plooij et~al.(2014)Plooij, Wolfslag, and Wisse}]{plooij2014open}
Plooij, M., Wolfslag, W., and Wisse, M. (2014).
\newblock Open loop stable control in repetitive manipulation tasks.
\newblock In \emph{2014 IEEE International Conference on Robotics and Automation (ICRA)}, 949--956. IEEE.

\bibitem[{Proctor et~al.(2016)Proctor, Brunton, and Kutz}]{proctor2016dynamic}
Proctor, J.L., Brunton, S.L., and Kutz, J.N. (2016).
\newblock Dynamic mode decomposition with control.
\newblock \emph{SIAM Journal on Applied Dynamical Systems}, 15(1), 142--161.

\bibitem[{Proctor et~al.(2018)Proctor, Brunton, and Kutz}]{proctor2018generalizing}
Proctor, J.L., Brunton, S.L., and Kutz, J.N. (2018).
\newblock Generalizing koopman theory to allow for inputs and control.
\newblock \emph{SIAM Journal on Applied Dynamical Systems}, 17(1), 909--930.

\bibitem[{Rahmani and Redkar(2024)}]{rahmani2024enhanced}
Rahmani, M. and Redkar, S. (2024).
\newblock Enhanced koopman operator-based robust data-driven control for 3 degree of freedom autonomous underwater vehicles: A novel approach.
\newblock \emph{Ocean Engineering}, 307, 118227.

\bibitem[{Schmid(2010)}]{schmid2010dynamic}
Schmid, P.J. (2010).
\newblock Dynamic mode decomposition of numerical and experimental data.
\newblock \emph{Journal of fluid mechanics}, 656, 5--28.

\bibitem[{Takens(1980)}]{takens2006detecting}
Takens, F. (1980).
\newblock Detecting strange attractors in turbulence.
\newblock In \emph{Dynamical Systems and Turbulence, Warwick 1980: proceedings of a symposium held at the University of Warwick 1979/80}, 366--381. Springer.

\bibitem[{Williams et~al.(2015)Williams, Kevrekidis, and Rowley}]{williams2015data}
Williams, M.O., Kevrekidis, I.G., and Rowley, C.W. (2015).
\newblock A data--driven approximation of the koopman operator: Extending dynamic mode decomposition.
\newblock \emph{Journal of Nonlinear Science}, 25(6), 1307--1346.

\bibitem[{Yang and Bhounsule()}]{yangkoopman}
Yang, C.M. and Bhounsule, P.A. (????).
\newblock Koopman operator based linear model predictive control for quadruped trotting.

\bibitem[{Yang and Bhounsule(2025)}]{yang2025koopman}
Yang, C.M. and Bhounsule, P.A. (2025).
\newblock Koopman operator based linear model predictive control for 2d quadruped trotting, bounding, and gait transition.
\newblock \emph{arXiv preprint arXiv:2507.14605}.

\end{thebibliography}

\end{document}